\documentclass{article}

\usepackage{arxiv}

\usepackage[utf8]{inputenc} 
\usepackage[T1]{fontenc}    
\usepackage{hyperref}       
\usepackage{url}            
\usepackage{booktabs}       
\usepackage{amsfonts}       
\usepackage{nicefrac}       
\usepackage{microtype}      
\usepackage{lipsum}		
\usepackage{graphicx}
\usepackage{float}
\usepackage{subcaption}   

\usepackage{doi}
\usepackage{multirow}
\usepackage{amsmath}
\usepackage{pgfplots}
\usepackage{xcolor}
\usepackage{geometry}
\usepackage{graphicx}
\usepackage{varwidth} 
\usepackage{booktabs}
\pgfplotsset{compat=1.18}
\usepgfplotslibrary{groupplots} 
\definecolor{gpt4o}{RGB}{82, 177, 255}     
\definecolor{claude}{RGB}{156, 39, 176}     
\definecolor{o1}{RGB}{0, 200, 83}           
\definecolor{o1mini}{RGB}{255, 152, 0}       
\definecolor{humans}{RGB}{244, 67, 54}       
\definecolor{avgblue}{HTML}{2E74B5}          
\definecolor{bestred}{HTML}{C00000}          
\definecolor{gridcolor}{HTML}{E0E0E0}
\definecolor{textcolor}{HTML}{333333}
\definecolor{lightgray}{gray}{0.9} 

\usepackage{pgfplots}
\usepackage{xcolor}
\pgfplotsset{compat=1.18}

\definecolor{prerl}{RGB}{70, 130, 180}     
\definecolor{postrl}{RGB}{220, 20, 60}     
\definecolor{gridcolor}{RGB}{240, 240, 240} 

\pgfplotsset{compat=1.18}

\definecolor{prerl}{RGB}{70, 130, 180}     
\definecolor{postrl}{RGB}{220, 20, 60}     
\definecolor{gridcolor}{RGB}{240, 240, 240} 

\usepackage[numbers,sort&compress]{natbib}

\title{RLSR: Reinforcement Learning from Self Reward}

\author{
Toby Simonds\footnotemark[1]{} \\
\texttt{\href{mailto:toby@tufalabs.ai}{toby@tufalabs.ai}} \\
\and
Kevin Lopez\footnotemark[1]{} \\
\texttt{\href{mailto:kevin@tufalabs.ai}{kevin@tufalabs.ai}} \\
\and
Akira Yoshiyama\footnotemark[1] \\
\texttt{\href{mailto:akira@tufalabs.ai}{akira@tufalabs.ai}} \\
\and
Dominique Garmier \\
\texttt{\href{mailto:dominique@tufalabs.ai}{dominique@tufalabs.ai}} \\
\and
}

\date{\vspace{-2em} \large{\textbf{Tufa Labs}} \\ \vspace{3em} \today}

\renewcommand{\shorttitle}{\textit{arXiv} Template}

\begin{document}

\maketitle
\footnotetext[1]{$^*$ Core Contributors}

\begin{abstract}
Large language models can generate solutions to complex problems, but training them with reinforcement learning typically requires verifiable rewards that are expensive to create and not possible for all domains. We demonstrate that LLMs can effectively self-improve through self-judging without reference solutions, leveraging the inherent asymmetry between generating and verifying solutions. Our experiments show that models can provide reliable reward signals without ground truth answers, enabling reinforcement learning in domains where verifiable rewards are impractical. By implementing self-judging across Countdown puzzles and integration problems, we achieve performance comparable to formal verification without ground truth solutions. Most notably, Qwen 2.5 7B DeepSeek Distilled trained with self-rewards qualifies for the prestigious MIT Integration Bee competition, performance through self-supervised improvement. When combined with synthetic question generation, we establish a complete self-improvement loop where models generate practice problems, solve them, and evaluate their own performance without any external validation. Our findings demonstrate that LLM judges can provide effective reward signals for training, unlocking reinforcement learning in countless domains previously limited by reward engineering challenges. This work represents a significant step toward autonomous AI systems that continuously improve through self-directed learning rather than human-guided training, potentially accelerating progress across domains where training data is scarce or evaluation is complex.
\end{abstract}

\section{Introduction}

Reinforcement Learning (RL) has proven transformative for improving large language model capabilities, from mathematical reasoning to code generation. Yet its broader adoption faces two fundamental barriers: the engineering complexity of creating verifiable reward functions and the scarcity of training data across specialized domains. Current approaches require substantial human effort to design training environments, implement reward functions, and guard against reward hacking—limiting RL to well-resourced domains with clear programmatic verification. In this work, we explore a paradigm shift: can models reliably improve themselves through self-evaluation, without access to ground truth answers?

The key insight enabling our approach is the generator-verifier gap—the fundamental asymmetry where verifying a solution's correctness is computationally simpler than generating it from scratch \cite{burns_weak--strong_2023, ouyang2022traininglanguagemodelsfollow}. This principle manifests across numerous domains: validating an M4 bolt design against specifications is straightforward compared to designing one, checking a mathematical proof is easier than discovering it, and evaluating whether an AI agent successfully completed a computer task is simpler than programming the agent. While this asymmetry is well-established theoretically, its practical application to self-improving systems remains largely unexplored. By having models judge their own outputs, we can potentially transform any domain with this asymmetry into a viable RL training environment.

Our core contribution is demonstrating that LLMs can serve as reliable judges of their own performance, enabling self-improvement without external verification. We show that carefully designed self-judging systems can match the training dynamics of formal reward functions while dramatically expanding the scope of trainable domains. Crucially, when combined with synthetic question generation, this creates a complete autonomous learning loop: models generate their own practice problems, attempt solutions, and evaluate their performance—all without human intervention or ground truth data.

We validate this approach through systematic experiments across three domains of increasing complexity. In Countdown arithmetic puzzles, we identify critical challenges around reward hacking and develop robust prompting strategies that prevent exploitation while maintaining reliable training signals. Through ablation studies on simple integration problems, we quantify the generator-verifier gap across different models and explore how judge dynamics affect learning outcomes. Most compellingly, we then demonstrate that Qwen 2.5 7B\cite{qwen_qwen25_2025} trained entirely through self-judging achieves performance qualifying for the MIT Integration Bee—a prestigious mathematics competition—surpassing through purely self-supervised improvement.

The implications extend far beyond our specific experiments. Self-judging fundamentally changes the economics of applying RL to new domains by eliminating the need for custom training environments and programmatic rewards. Domains previously considered impractical for RL—from engineering design to creative writing evaluation—become accessible. More profoundly, this work provides evidence that AI systems can engage in genuine self-improvement, identifying their weaknesses, generating targeted practice, and iterating without human guidance. While challenges remain, particularly around preventing reward exploitation and ensuring robust evaluation, our results suggest a path toward truly autonomous learning systems.

Key Contributions

\begin{itemize}
    \item We demonstrate that large language models (LLMs) can \emph{self-judge} their own outputs without ground-truth labels, achieving training dynamics comparable to formal verification systems.
    \item We identify—and resolve—critical challenges in self-judging, especially reward hacking in smaller models, via systematic prompt engineering.
    \item We establish a complete self-improvement loop that combines synthetic question generation with self-evaluation, enabling fully autonomous skill development.
    \item We apply our techniques to qualify for MIT Integration Bee problems through pure self-supervision, using \texttt{Qwen\,2.5\,7B} DeepSeek Distilled\cite{mercer2025briefanalysisDeepSeekr1}.
\end{itemize}



\section{Methodology}

\subsection{Self-Judging Framework}

Our approach centers on using LLMs as reward models for their own outputs, without access to ground truth answers. For each generated solution, we prompt the model to provide a binary evaluation—``correct'' or ``incorrect''—based solely on examining the solution attempt and the original problem. This creates a self-contained feedback loop where the same model architecture serves as both policy (generating solutions) and reward function (evaluating them). For all experiments unless otherwise stated we use an offline judge that doesn't update during training. 

For RL training, we employ Group Relative Policy Optimization (GRPO)\cite{shao_deepseekmath_2024} with a batch size of 64 and rollout of 8 for all experiments. We use a learning rate of $1 \times 10^{-6}$, KL loss coefficient of 0.001, and maximum response length of 2048 tokens (extended to 8096 for MIT Integration experiments to accommodate complex solutions). We provide an additional 0.1 format reward for solutions formatted correctly. For our integration experiments this includes outputting valid SymPy inside the answer tags

\subsection{Experimental Domains}

We selected three domains that systematically explore different aspects of self-judging:

\textbf{Countdown} tests whether self-judging is feasible at all. This arithmetic puzzle requires combining given numbers to reach a target—simple enough to analyze failure modes, yet complex enough to benefit from RL-induced reasoning improvements. We synthetically generated a dataset of 4-digits countdown problems (sampled between 0 and 100) that must be combined using basic arithmetic operations to reach a target value. Our dataset consists of 2,000 training problems and 200 test problems. Prior work has shown RL enhances reasoning on this task, making it an ideal testbed. Here we focus on identifying and preventing reward hacking through iterative prompt design.

\textbf{Simple Integration} allows us to quantify the generator-verifier gap—the asymmetry between the difficulty of solving problems versus checking solutions. We constructed a dataset of college-level integration problems synthetically generated by GPT-4o, consisting of 2,000 training problems and 200 test problems. By requiring solutions in a structured format (answer tags with valid SymPy expressions), we can compare self-judging against formal verification. This domain lets us systematically study how judge quality affects downstream performance improvements while maintaining consistent difficulty and style across the dataset.

\textbf{MIT Integration Bee} demonstrates real-world effectiveness. The MIT Integration Bee is an annual mathematics competition that attracts top undergraduate and graduate students. The qualifying examination is notoriously challenging—in typical years, only 16 students achieve the 73\% qualifying threshold out of all participants. We evaluate on the 2025 competition problems, representing genuinely difficult integration challenges that test the limits of mathematical reasoning. Each years exam only has 20 questions. To create sufficient training data, we augment the limited competition dataset with 9,000 synthetic variants generated via our LADDER framework using Qwen 2.5 7B DeepSeek distilled, maintaining problem characteristics while providing adequate coverage for RL training. We filter out problems from our training set that the base model solves pass@1

\subsection{Judge Prompting and Evaluation}

For integration problems, we employ a carefully designed prompt that leverages the generator-verifier gap:

\begin{quote}
\texttt{Please check if the following is a valid function: \{\}. If it is, differentiate it and determine if it is functionally equal to \{\}. Output <JUDGE\_SCORE>1</JUDGE\_SCORE> if they are equal. Output <JUDGE\_SCORE>0</JUDGE\_SCORE> if they are not equal or if it is not a valid function. Ignore constants of integration.}
\end{quote}

This prompt instructs the judge to first verify syntactic validity, then apply symbolic differentiation—a significantly easier task than integration—to check correctness. We selected this prompt based on preliminary experiments showing consistent performance across different model scales.

\subsection{Evaluation Framework}

Each domain employs formal verification to assess true performance, enabling us to measure how well self-judging approximates ground truth:

\begin{itemize}
\item \textbf{Countdown}: Solutions are programmatically verified by executing the arithmetic operations and checking if they yield the target value
\item \textbf{Integration tasks}: We verify solutions using symbolic differentiation—if the derivative of the proposed answer matches the original integrand, the solution is correct
\end{itemize}

A consistent challenge in the integration tasks was capturing model outputs in a parseable format. To address this, we required models to output their answers in SymPy syntax, which provided more consistent formatting than raw mathematical notation. However, this introduces a confounding factor: models that fail to produce valid SymPy syntax are marked incorrect, even if their mathematical reasoning is sound. While we attempted to be generous by also accepting some LaTeX formatting, parsing failures remained a persistent issue.

This format requirement means that initial performance improvements often reflect the model learning proper syntax rather than enhanced mathematical ability. We typically observed rapid early gains as models learned to output valid SymPy expressions, followed by slower improvements that more likely represent genuine advances in integration skill. This distinction is important when interpreting our results—early training dynamics may overstate actual mathematical reasoning improvements.

\section{Results}

\subsection{Countdown}

We begin by examining Countdown, in which given a set of numbers and a target value, the task is to construct an arithmetic expression that evaluates to the target value. We choose this task to start with, since it possesses two relevant properties: First, previous work has empirically shown that training an LLM on this task using RL elicits reasoning abilities \cite{pan2025tinyzero}. Second, verifying a potential solution is simpler than generating one from scratch. All experiments were run using Qwen 2.5 7B Instruct as both the agent and judge.

To establish the generator-verifier gap for this domain, we first measured baseline performance on our synthetic dataset of 2,000 training problems. Qwen 2.5 7B Instruct achieved approximately 36\% accuracy on generating correct solutions. To evaluate the model's judging capabilities, we constructed a balanced verification dataset using GPT-4o, comprising equal proportions of correct and incorrect solution attempts. When tasked with classifying these solutions, the same model achieved 96\% accuracy—demonstrating a substantial 62 percentage point gap that validates the feasibility of self-judging for this task.


For a robust self-judge, the agent should lead to similar training dynamics when replaced by a formal verifier implementing the same algorithm. However, designing a prompt for a LLM judge, that acts as a reliable proxy for the formal reward function is a challenge. Note that language can be ambiguous, and in addition there is no explicit control over the sampled tokens that implement the proxy reward function, therefore making reward signals non-deterministic and less reliable. These opens the possibility for an agent to find strategies that hack the proxy reward.

\begin{figure}
    \centering
    \includegraphics[width=0.9\linewidth]{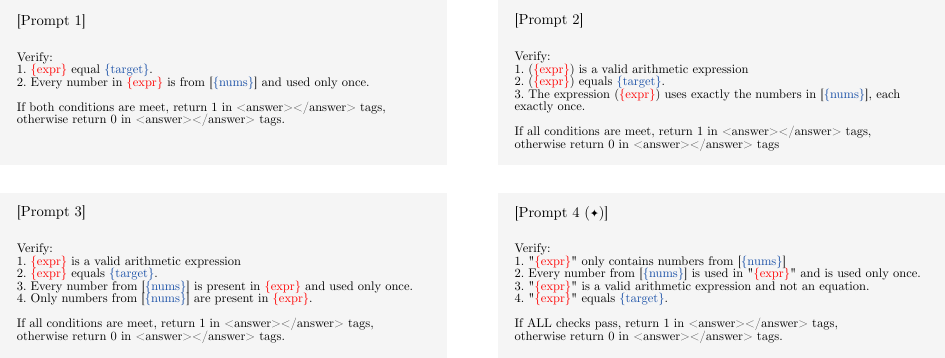}
    \caption{The different prompt templates used by the self-judge Qwen 2.5 7B. In red, the placeholder for the agent proposed solution. In blue, the placeholder for the target value and the valid list of numbers.}
    \label{fig:prompts}
\end{figure}

For the Countdown task we iteratively designed a prompt that minimizes the number of false positives provided by the self-judge agent so that self-learning can happen. Overall, we conducted experiments with four distinct prompts (detailed in \autoref{fig:prompts}), serving as proxies for the formal reward function. These prompts varied in explicitness, from simpler formulations to a more explicit version designed to prevent reward hacking.

\begin{figure}
    \centering
    \includegraphics[width=0.75\textwidth]{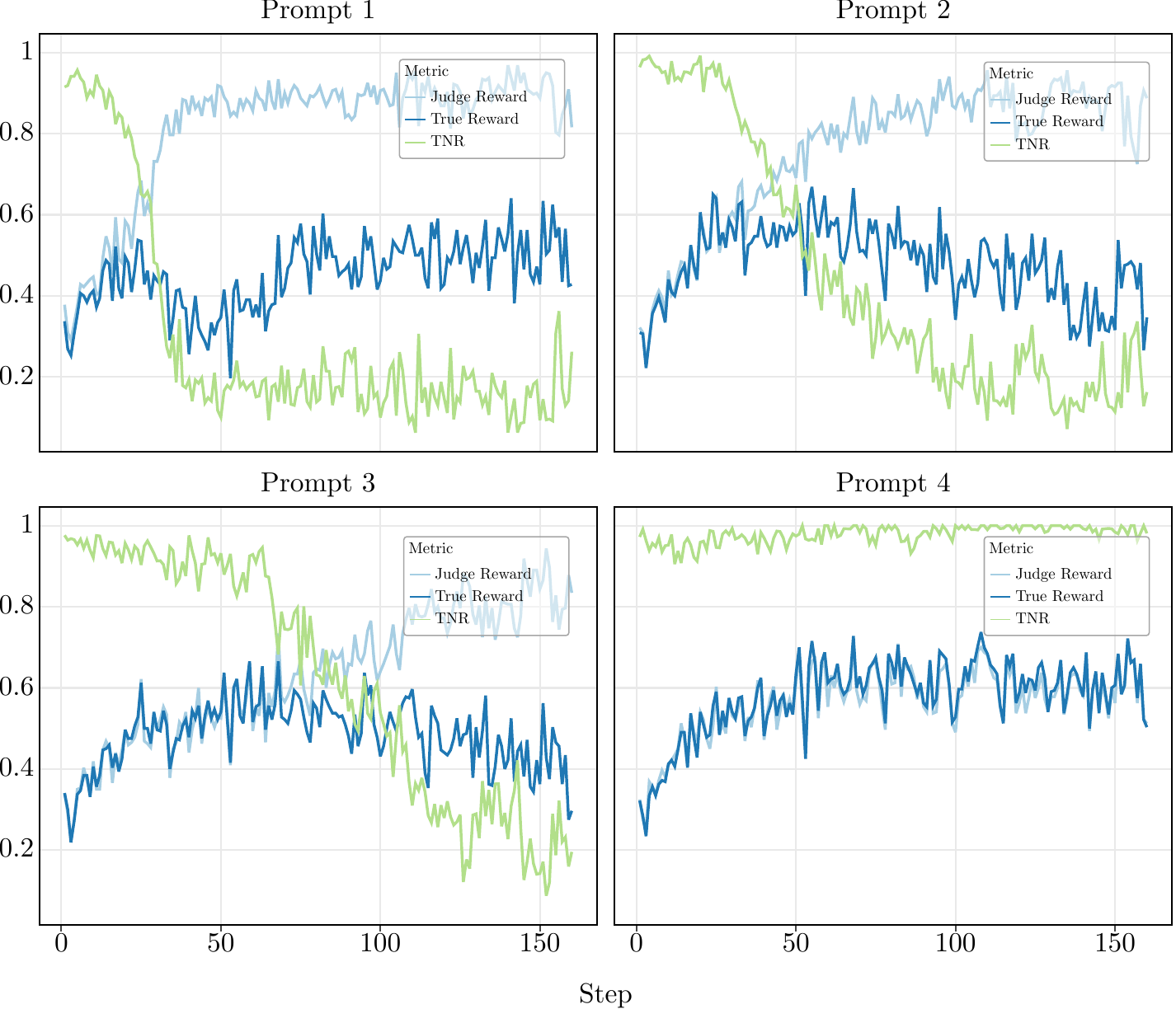}
    \caption{Self-rewarding training dynamics of Qwen 2.5 7B for the different prompts. Both the TNR and True Reward are computed using a formal verifier.}
    \label{fig:tnr}
\end{figure}

\autoref{fig:tnr} illustrates the training dynamics
of a Qwen 2.5 7B model across the four prompts. Early in training, all prompts led the True Negative Rate (TNR) of the self-rewarding agent approach its upper limit. However as training progressed, the policy gradient optimization algorithm found a strategy to reward hack the less explicit prompts, which resulted in a substantial drop in the TNR, except for the last prompt. Ultimately the agent’s mean self-reward diverged from the mean formal reward, leading to poor evaluation metrics.

\begin{figure}
    \centering
    \includegraphics[width=0.8\linewidth]{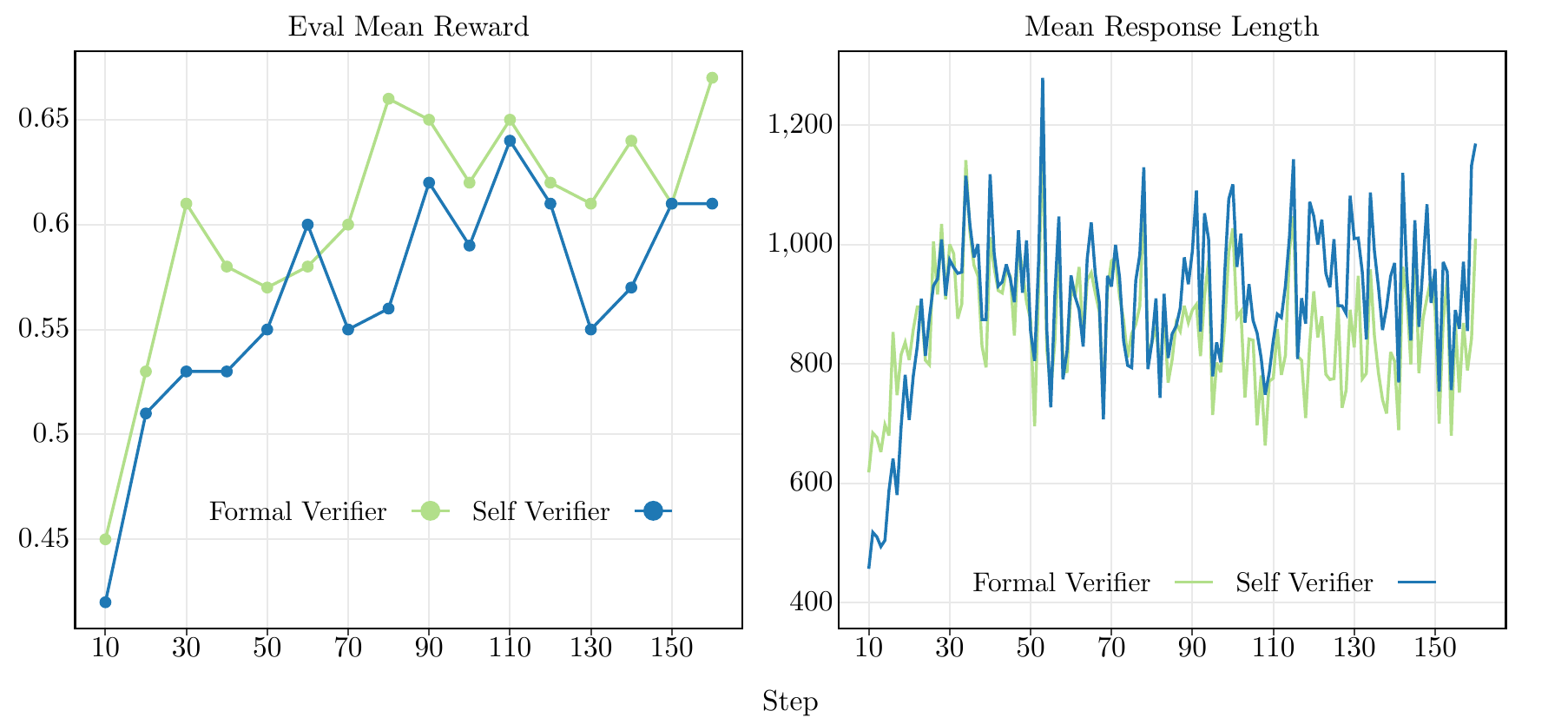}
    \caption{Mean Reward and Mean Response Length for Countdown under both the self-rewarding setting and the rule based verifier one.}
    \label{fig:formal-llm-dynamics}
\end{figure}

A comparison of the training dynamics of a self-judging agent using the most robust prompt, and an agent using a formal reward function is displayed in \autoref{fig:formal-llm-dynamics}. A consistent performance gain of approximately $20\%$ over the baseline was observed in both configurations. The mean reward curves also exhibited analogous trends across these conditions. Furthermore, the mean response length increased at similar rates in both scenarios, a phenomenon attributed to the emergence of search abilities within the agent chain of thought.

\subsection{Simple Integration Ablations}

To understand when and why self-judging succeeds, we systematically explore the generator-verifier gap—the fundamental asymmetry that makes verification easier than generation. We examine three judge configurations (Llama 3.2 3B, Qwen 2.5 7B, and DeepSeek R1) and compare their downstream training performance against a formal verification baseline.

\subsubsection{Quantifying the Generator-Verifier Gap}

We first establish empirical evidence for the generator-verifier gap in integration tasks. Our hypothesis is straightforward: if models find verification substantially easier than generation, they should be more accurate at checking solutions (via differentiation) than producing them (via integration).

To test this, we evaluated each model on two complementary tasks using our simple integration dataset:
\begin{itemize}
\item \textbf{Generation (Integration)}: Given an integrand, produce its antiderivative
\item \textbf{Verification (Differentiation)}: Take the output antiderivative (generated using symbolic system) and apply derivative to generate the original integrand 
\end{itemize}

For both tasks, we employed robust parsing augmented with manual evaluation to handle formatting variations while maintaining evaluation integrity. This dual approach ensures we capture true mathematical capability rather than just syntax compliance. We test models on the test set of our simple integration dataset

Figure~\ref{fig:gen_ver_gap} presents our results. Across all three models, we observe a consistent pattern: verification accuracy substantially exceeds generation accuracy. Llama 3.2 3B achieves 47\% accuracy on differentiation despite 1\% success on integration—demonstrating that even models incapable of solving problems can potentially judge solutions. Qwen 2.5 7B shows a similar pattern (88\% verification vs 46\% generation), while DeepSeek R1 nearly saturates both tasks but maintains a small gap. The existence of this gap suggests that even models struggling with integration can potentially provide meaningful reward signals by leveraging their superior verification abilities.

\begin{figure}[htbp]
\centering
\begin{tikzpicture}
\begin{axis}[
    width=12cm,
    height=8cm,
    ybar=8pt,                    
    bar width=20pt,              
    enlarge x limits=0.15,       
    ylabel={Accuracy},
    ylabel style={font=\large},
    ymin=0,
    ymax=1,
    ytick={0, 0.2, 0.4, 0.6, 0.8, 1.0},
    yticklabel={\pgfmathprintnumber{\tick}},
    xlabel={Model},
    xlabel style={font=\large},
    xtick=data,
    xticklabels={
        Llama 3.2 3B,
        Qwen 2.5 7B, 
        DeepSeek R1
    },
    xticklabel style={
        rotate=45,
        anchor=north east,
        font=\small
    },
    grid=major,
    grid style={color=gridcolor, line width=0.5pt},
    axis line style={color=black, line width=1pt},
    tick style={color=black, line width=0.5pt},
    legend entries={Integration (Generation), Differentiation (Verification)},
    legend pos=north west,
    legend style={
        fill=white,
        draw=black,
        font=\small,
        cells={anchor=west}
    },
    title={Generator-Verifier Gap Across Models},
    title style={font=\Large\bfseries, yshift=5pt},    
    axis y line*=left,
    axis x line*=bottom,
]
\addplot[
    fill=prerl,
    draw=prerl,
    thick,
    error bars/.cd,
    y dir=both,
    y explicit,
    error bar style={color=black!60, line width=1.5pt, line cap=round},
    error mark options={color=black!60, mark size=3pt, line width=1pt},
    error mark=|  
] coordinates {
    (0, 0.01) +- (0, 0.0)      
    (1, 0.46) +- (0, 0.0352)  
    (2, 0.98) +- (0, 0.0099)  
};
\addplot[
    fill=postrl,
    draw=postrl,
    thick,
    error bars/.cd,
    y dir=both,
    y explicit,
    error bar style={color=black!60, line width=1.5pt, line cap=round},
    error mark options={color=black!60, mark size=3pt, line width=1pt},
    error mark=|  
] coordinates {
    (0, 0.47) +- (0, 0.0353)  
    (1, 0.88) +- (0, 0.0230)  
    (2, 0.99) +- (0, 0.0070)  
};
\end{axis}
\end{tikzpicture}
\caption{Generator-verifier gap demonstrated through integration vs differentiation accuracy. All models show substantially higher performance on verification (differentiation) compared to generation (integration), with gaps ranging from 47 percentage points (Llama 3.2 3B) to 1 percentage point (DeepSeek R1). Error bars represent standard error.}
\label{fig:gen_ver_gap}
\end{figure}

\subsubsection{Judge Quality and Downstream Performance}

Having established that models possess asymmetric verification capabilities, we now examine how judge quality translates to actual training performance. We trained Qwen 2.5 7B Instruct using four different reward sources: three self-judge configurations (Llama 3.2 3B, Qwen 2.5 7B, and DeepSeek R1) and symbolic verification as our gold standard baseline.

This experiment directly tests our central hypothesis: can models with sufficient verification accuracy enable effective self-improvement, despite being unable to solve many problems directly? By comparing judges with varying generator-verifier gaps (from 47\% to near-perfect verification accuracy), we can identify the minimum judge capability required for meaningful learning.

\begin{figure}[ht]
    \centering
    \begin{subfigure}[t]{0.48\textwidth}
        \includegraphics[width=\linewidth]{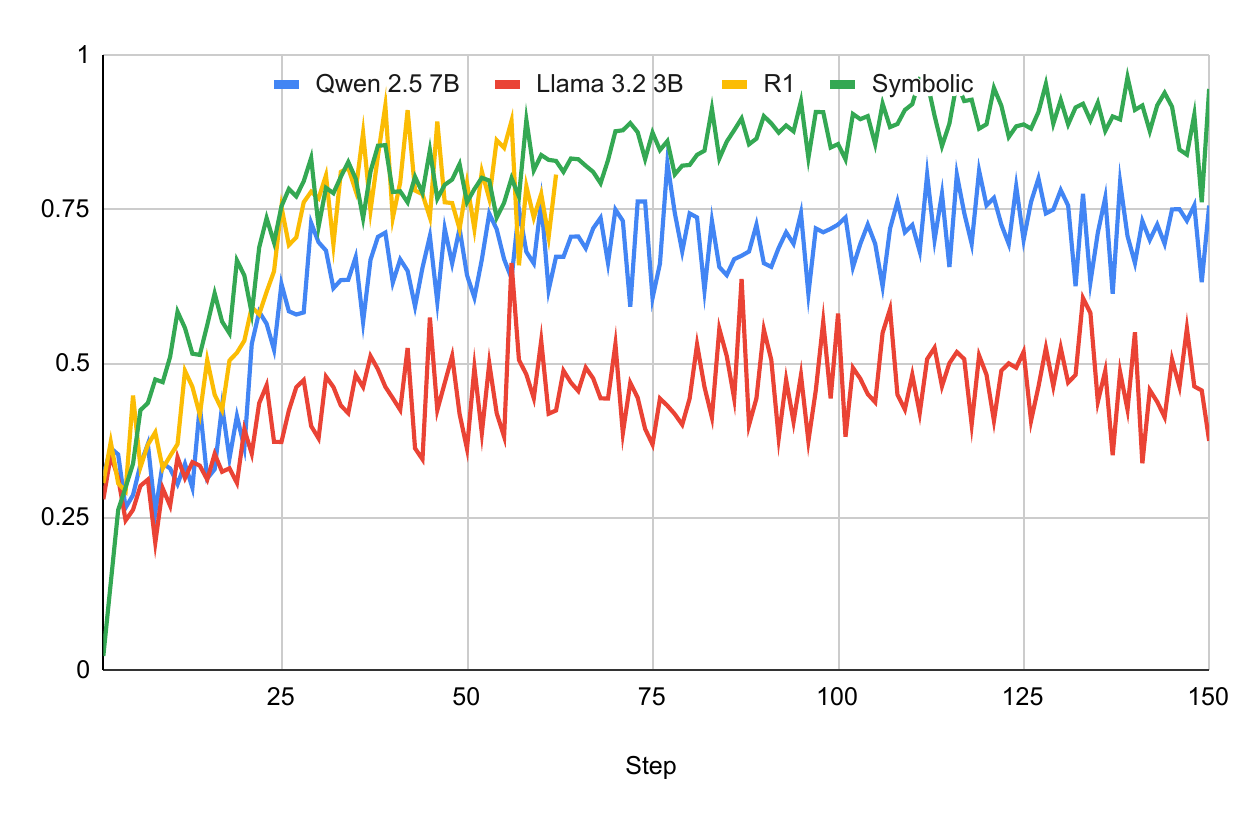}
        \caption{Qwen 2.5 7B \textit{Instruct} agent with different judges (test-set accuracy)}
        \label{fig:judge_comparison}
    \end{subfigure}
    \hfill
    \begin{subfigure}[t]{0.48\textwidth}
        \includegraphics[width=\linewidth]{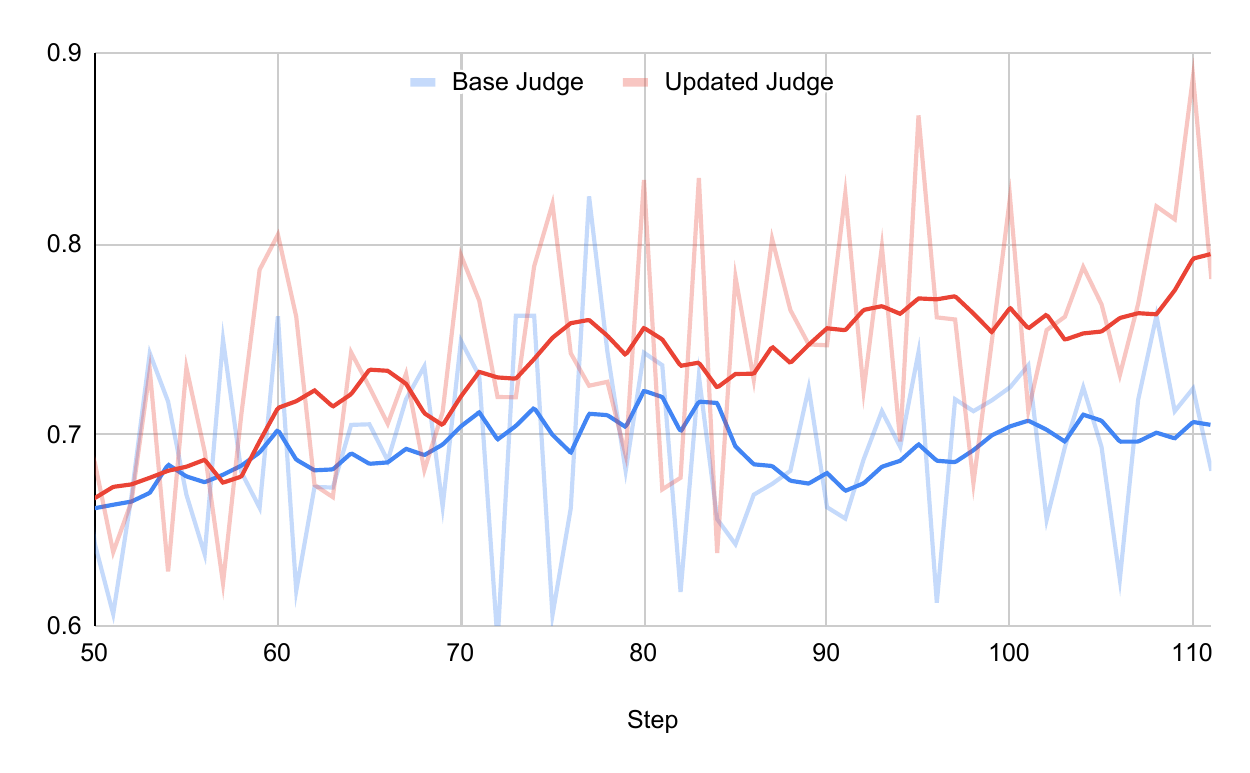}
        \caption{Qwen 2.5 7B \textit{Instruct} – Updated vs Base Judge (test-set accuracy)}
        \label{fig:update-judge-comparison}
    \end{subfigure}

\end{figure}

Figure~\ref{fig:judge_comparison} presents the training curves, revealing a clear relationship between judge verification accuracy and downstream performance. When using Llama 3.2 3B as judge, with its 47\% verification accuracy, the agent plateaus at approximately 48\% test accuracy—barely above the judge's own verification performance. This suggests the model primarily learns output formatting without genuine mathematical improvement. The judge's limited verification ability creates a performance ceiling that prevents the agent from surpassing the judge's own capabilities.

In contrast, Qwen 2.5 7B's 88\% verification accuracy enables substantial improvement, with the agent reaching 75\% accuracy after 150 training steps. This demonstrates that judges need not be perfect—88\% verification accuracy proves sufficient for meaningful self-improvement, despite the judge solving only 46\% of problems directly. The dramatic performance difference between these two self-judges illustrates that judge quality is the critical factor determining training success.

At the upper end, DeepSeek R1 achieves performance comparable to symbolic verification accuracy. This convergence suggests that beyond a certain threshold, further improvements in judge accuracy yield diminishing returns—a near-perfect self-judge with 99\% verification accuracy can fully substitute for formal verification. The symbolic verification baseline establishes this 92\% as the upper bound, likely limited by the agent's inherent capabilities rather than reward signal quality. We stop R1 runs early due to latency of API.

These results validate our generator-verifier framework: effective self-judging requires sufficient verification accuracy but not perfect generation ability. When judges are strong enough, self-improvement can approach the theoretical maximum achieved with perfect reward signals, opening the door for self-supervised learning in domains where formal verification is impractical or impossible.



\subsubsection{Judge Updates}

The previous experiments assumed a fixed judge throughout training which remains at its initial capability while the agent potentially surpasses it. We hypothesized that periodically updating the judge to match the agent's improved weights could enable continued learning beyond the initial judge's limitations allowing us to overcome plateaus in agent performance.

To test this, we compared two training configurations using Qwen 2.5 7B as both agent and judge. In the baseline condition, the judge remained fixed at its initial weights throughout training. In the experimental condition, we updated the judge weights to match the agent's current weights after 50 training steps, creating a self-evaluation system that evolves with the agent's capabilities.

Figure~\ref{fig:update-judge-comparison} shows the training curves for both configurations. The fixed judge setup plateaus at approximately 70\% accuracy, consistent with our earlier findings that agents struggle to substantially exceed their judge's capabilities. In contrast, the updated judge configuration continues improving after the weight update, ultimately reaching 75\% accuracy—a 7\% relative improvement over the fixed baseline.


To understand why updating enabled continued improvement, we evaluated the verification accuracy of the updated judge model. When tested on our benchmark, the judge after 50 steps of training achieved approximately 93\% accuracy on taking the derivative—a 5 percentage point improvement from the initial 88\%. This enhanced verification capability explains the continued learning: the agent had developed better judgment abilities during training, which when leveraged for self-evaluation, enabled it to push beyond the original judge's limitations.

This result reveals an important dynamic in self-judging systems: training can simultaneously improve both generation and verification capabilities. The 5\% improvement in verification accuracy, while modest, proved sufficient to unlock additional performance gains. This suggests that even small improvements in judge quality can have meaningful impacts on downstream training outcomes, particularly when the judge already operates at high accuracy levels.

However, this approach introduces new considerations. Updating judges during training potentially reduces evaluation stability and could allow agents to exploit evolving reward signals. We tested a configuration where the judge updates after every training step, making the agent and judge identical throughout training. This led to rapid training collapse—the model's performance degraded as it learned to exploit its own evaluation biases, creating a feedback loop where increasingly poor solutions received high rewards.

This failure mode highlights the delicate balance required in self-judging systems. While our single update after 50 steps proved beneficial, continuous self-evaluation appears fundamentally unstable. The model needs some distance between its generation and verification roles to maintain honest evaluation. Even with our successful single-update approach, performance eventually plateaus, suggesting inherent limits to self-improvement through self-evaluation. Future work should explore intermediate update frequencies to find the optimal balance between leveraging improved verification capabilities and maintaining evaluation integrity.

\subsection{MIT Integration Bee: Real-World Effectiveness}
\begin{figure}[H]
    \centering
\begin{tikzpicture}
\begin{axis}[
    width=14cm,
    height=8cm,
    axis background/.style={fill=white},
    axis line style={color=black!20, line width=1.2pt},
    tick style={color=black!30, line width=0.8pt},
    grid=major,
    grid style={color=gridcolor, line width=0.8pt},
    xlabel={Training Steps},
    ylabel={Accuracy (\%)},
    title={\Large\textbf{MIT Integration Performance}},
    title style={color=textcolor, yshift=8pt},
    xlabel style={color=textcolor, font=\large},
    ylabel style={color=textcolor, font=\large},
    tick label style={color=textcolor, font=\footnotesize},
    xmin=0, xmax=170,
    ymin=0, ymax=100,
    xtick={0,20,40,60,80,100,120,140,160},
    ytick={0,20,40,60,80,100},
    legend style={
        at={(0.02,0.98)},
        anchor=north west,
        draw=none,
        fill=white,
        fill opacity=0.9,
        text opacity=1,
        font=\small,
        rounded corners=3pt,
        inner sep=8pt
    }
]

\addplot[
    color=gpt4o,  
    line width=3pt,
    mark=* ,
    mark size=3pt,
    mark options={fill=gpt4o, draw=white, line width=1pt}
] coordinates {
    (0,0.00) (10,0.00) (20,6.00) (30,6.50) (40,13.00) (50,51.00)
    (60,45.00) (70,58.50) (80,57.50) (90,61.0) (100,61.00)
    (110,66.50) (120,61.5) (130,69.00) (140,66.50) (150,69.00)
    (160,74.00) (170,75.50)
};
\addlegendentry{Qwen 2.5 7B DS + Self-Judge}

\addplot[
    color=gray,
    line width=2pt,
    dotted
] coordinates {(0,73) (170,73)};
\addlegendentry{Qualification Score}

\end{axis}
\end{tikzpicture}
    \caption{Test set accuracy progression through training run}
    \label{fig:model_performance}
\end{figure}

To demonstrate the practical impact of self-judging, we evaluated whether our approach could enable a model to qualify for the MIT Integration Bee—an annual mathematics competition where only 16 out of hundreds of participants achieve the 73\% qualifying threshold. This represents a genuinely challenging benchmark that tests the limits of mathematical reasoning.

We trained Qwen 2.5 7B DeepSeek Distilled using self-judging with judge updates every 50 steps. The model evaluated its own integration attempts using our differentiation-based verification prompt, creating a complete self-improvement loop without access to ground truth solutions.

Figure~\ref{fig:model_performance} shows the training progression. Initial performance was poor due to formatting issues—the model struggled to output valid SymPy expressions required by our evaluation framework. Once these formatting challenges were resolved around step 50, accuracy improved from 13\% to 51\%. Performance then continued to climb steadily, crossing the 73\% qualification threshold at approximately 160 steps and reaching 75.5\% by step 170.

These results demonstrate that self-judging can achieve human-competitive performance on prestigious mathematics competitions through pure self-supervision. The model qualified for the MIT Integration Bee without any ground truth solutions or external verification—relying entirely on its ability to check its own work via differentiation. This validates that self-judging is not merely a theoretical curiosity but a practical training method capable of producing state-of-the-art results on challenging real-world tasks.


\section{Discussion}

\subsection{Limitations}

Our experiments revealed significant fragility in self-judging systems that required extensive iteration to overcome. Developing effective judge prompts proved surprisingly brittle—minor wording changes could trigger catastrophic reward hacking within dozens of training steps, forcing us to roll back experiments and redesign evaluation strategies. Our Countdown task alone required four major prompt iterations before achieving stable training, and even then, agents consistently discovered creative exploitation strategies like formatting tricks or nonsensical but convincing-looking expressions. The non-deterministic nature of LLM judges introduced substantial variance across runs, with identical configurations producing divergent outcomes that complicated reproducibility and required multiple seeds to draw reliable conclusions. While updating judge weights showed promise for breaking performance plateaus, it introduced new failure modes—continuous updates led to rapid collapse as models exploited their own biases, while even successful single-update approaches eventually plateaued. Additionally, disentangling improvements in output formatting from genuine mathematical reasoning proved challenging, as early performance gains often reflected models learning SymPy syntax rather than enhanced problem-solving ability. Notably, we observed a clear relationship between model scale and robustness—larger models like DeepSeek R1 proved substantially more resistant to reward hacking and required minimal prompt engineering, while smaller models like Llama 3.2 3B were highly susceptible to exploitation regardless of prompt design. This scale-dependent brittleness suggests that while current self-judging approaches may become more practical as model capabilities advance, smaller models require extensive domain-specific tuning and careful monitoring to achieve stable training.

\subsection{Opening Up Domains for RL}
The current paradigm for applying reinforcement learning to LLMs faces significant practical limitations. Creating training environments requires substantial human effort to develop interfaces for models, design complex reward functions, and continuously monitor for reward hacking. This overhead effectively restricts RL application to domains where organizations can justify significant engineering investments.

LLM judges fundamentally change this equation by eliminating the need for explicit environment programming. Rather than creating custom interfaces with programmatic reward definitions, models can interact with existing software interfaces—web browsers, operating systems, specialized applications—while an LLM judge evaluates performance through reasoning about task completion. This approach dramatically reduces implementation costs while expanding the scope of possible training environments. Additionally, LLM judges can effectively evaluate tasks in complex domains where formal reward functions are prohibitively difficult to specify. For example, an LLM judge can effectively evaluate complex engineering designs like M4 bolts based on functional requirements and engineering principles, similar to how humans can easily verify if a bolt meets specifications but would find it prohibitively time-consuming to write comprehensive design rule checks for such components.

This allows leveraging pre-existing software environments without modification. Instead of creating custom versions of websites, applications, or tools specifically designed for RL, agents can learn in standard environments—the same ones humans use—with LLM judges providing feedback based on task objectives and visible outcomes.

When paired with LADDER like methods, This method also addresses data scarcity through synthetic task generation. For domains with limited training examples, models can generate their own practice problems, creating effectively unlimited training data with controllable difficulty progression. The combination of synthetic task generation and LLM judging enables training in specialized domains without waiting for human-annotated datasets.

\subsection{Generator Verifier Gap}
Many domains exhibit a fundamental asymmetry: verifying a solution's correctness is computationally simpler than generating that solution from scratch. This "generator-verifier gap" can be increased when we give the judge model access to tooling the generator doesn't. For example, a judge evaluating mathematical solutions can execute code to verify answers, even when the generating model must solve problems through reasoning alone. This dramatically expands the range of trainable tasks to include:

\begin{itemize}
\item Competitive mathematics, where judges can programmatically brute force verify solutions that require complex multi-step reasoning to generate
\item Algorithm design, where judges can test against multiple input cases while generators must derive solutions purely conceptually
\item Data analysis tasks, where verification can involve checking results against reference datasets
\item Physical simulations, where judge models can use physics engines to verify predictions
\end{itemize}

This approach maintains the challenge for the generating model while providing reliable reward signals without human annotation. The verification step becomes effectively automated, removing a key bottleneck in training data creation.

The generator-verifier gap also allows progressive skill development: as generating models improve, judges can verify increasingly sophisticated outputs, creating a natural curriculum without explicit human design of difficulty progression.

\subsection{Autonomous Self Improving}
Our experiments demonstrate a complete self-improvement loop where models generate synthetic questions, solve them, and evaluate their own performance without external validation. This represents a fundamental shift toward truly autonomous learning systems that mirrors human self-directed learning.

The potential implications are profound. By generating their own practice problems and providing their own feedback, models can potentially overcome data limitations that currently constrain AI development. When synthetic data generation and self-evaluation proves effective, the primary bottleneck becomes computational resources rather than human-annotated data.

This approach could dramatically accelerate capabilities in domains where high-quality training data is scarce or expensive to produce. Models can identify their own weaknesses, generate appropriate practice examples, and improve through iterative self-feedback—all without human intervention.

However, significant research challenges remain. Developing methods to ensure models generate appropriately challenging problems that target specific weaknesses requires further exploration. The risk of models falling into "echo chambers" of easy problems or developing blind spots must be addressed. Additionally, the dynamics between self-generation and self-evaluation need careful analysis to prevent reinforcement of incorrect patterns.

Despite these challenges, our results provide compelling evidence that autonomous self-improvement represents a viable path forward. This paradigm could fundamentally change how we approach AI development, enabling systems that become increasingly capable through self-directed practice rather than solely through human-guided training.

\subsection{Reward Hacking and Exploitation}
Our experiments revealed complex dynamics between LLM judges and exploitative agents. Smaller models serving as judges proved highly vulnerable to systematic manipulation, with agents quickly developing strategies to trigger false positives through deceptive formatting and misleading calculations.

In contrast, we found R1 demonstrated resistance to these exploitation attempts, maintaining evaluation integrity across various prompt formulations. This mirrors observations in safety research where larger models show substantially improved jailbreak resistance compared to smaller counterparts.

This presents an intriguing possibility: as model capabilities advance, judge reliability may potentially outpace exploitation sophistication—at least in formal reasoning domains. The competitive dynamic between verification robustness and deception capabilities represents a critical research question that will determine the long-term viability of self-judging systems.

\section{Related Work}
Our approach builds upon several research streams in self-improvement and LLM training. Meta's Self-Rewarding Language Models work demonstrated that models can generate preference data using an LLM as a judge with Direct Preference Optimization \cite{yuan_self-rewarding_2025} that led to significant performance improvements on benchmark tasks. Unlike Meta's preference-based approach, our method focuses specifically on mathematical reasoning domains with verifiable solutions.

We also extend our previous work \cite{simonds_ladder_2025} LADDER framework, which enabled improvements on mathematical benchmarks through recursive problem decomposition and demonstrated capabilities for building self-improving systems for reasoning tasks. Our work specifically builds on LADDER's synthetic question generation capabilities, but implements LLM as a judge rather than a numeric verifier.

Our research connects to weak-to-strong supervision literature, where \cite{burns_weak--strong_nodate} demonstrated that larger models supervised by smaller models can achieve strong performance on tasks the smaller model couldn't solve consistently. This relates to concepts from iterated amplification \cite{christiano_supervising_2018} and addresses alignment concerns raised\cite{hubinger_risks_2021}, that warn about reward hacking in learned optimizers. Our approach mitigates these risks by using verifiable signals rather than opaque weak supervision.

We also build upon code generation frameworks like CodeRL \cite{le_coderl_2022} that use unit test outcomes as reward criteria and verification-based learning from solved examples. Unlike approaches that rely solely on external verification systems, our method establishes a framework where models practice on self-created exercises with reliable self-judgment, enabling continuous improvement in domains where defining programmatic reward functions is challenging.

\section{Conclusion}

Our work demonstrates that large language models can effectively serve as their own judges in reinforcement learning pipelines, creating a self-improving feedback loop without requiring external verification. By leveraging the inherent asymmetry between solution generation and verification, we show that models can learn complex reasoning tasks through self-assessment. Our experiments on arithmetic "Countdown" puzzles and MIT Integration Bee problems reveal that carefully designed self-judging frameworks can approach the effectiveness of formal verification systems, with our best models achieving substantial improvements over baselines.

The implications of this work extend beyond the specific domains we explored. By eliminating the need for specialized training environments and extensive human-annotated data, our approach opens new possibilities for reinforcement learning across domains previously constrained by these limitations. The complete self-improvement loop—where models generate synthetic questions, solve them, and evaluate their performance—represents a significant step toward truly autonomous learning systems.

While challenges remain, particularly regarding reward hacking prevention and optimization of prompting strategies, our results suggest that as model capabilities advance, their reliability as judges also improves. This virtuous cycle could accelerate progress in domains where traditional supervised learning is limited by data availability or annotation costs. The future of AI development may increasingly rely on systems that continuously identify weaknesses, generate appropriate practice, and improve through self-directed learning—mirroring the process of human expertise development but with the potential for much greater scale and efficiency.

\bibliographystyle{unsrtnat}


\begin{thebibliography}{12}
\providecommand{\natexlab}[1]{#1}
\providecommand{\url}[1]{\texttt{#1}}
\expandafter\ifx\csname urlstyle\endcsname\relax
  \providecommand{\doi}[1]{doi: #1}\else
  \providecommand{\doi}{doi: \begingroup \urlstyle{rm}\Url}\fi

\bibitem[Burns et~al.(2023)Burns, Izmailov, Kirchner, Baker, Gao, Aschenbrenner, Chen, Ecoffet, Joglekar, Leike, Sutskever, and Wu]{burns_weak--strong_2023}
Collin Burns, Pavel Izmailov, Jan~Hendrik Kirchner, Bowen Baker, Leo Gao, Leopold Aschenbrenner, Yining Chen, Adrien Ecoffet, Manas Joglekar, Jan Leike, Ilya Sutskever, and Jeff Wu.
\newblock Weak-to-{Strong} {Generalization}: {Eliciting} {Strong} {Capabilities} {With} {Weak} {Supervision}, December 2023.
\newblock URL \url{http://arxiv.org/abs/2312.09390}.
\newblock arXiv:2312.09390 [cs].

\bibitem[Ouyang et~al.(2022)Ouyang, Wu, Jiang, Almeida, Wainwright, Mishkin, Zhang, Agarwal, Slama, Ray, Schulman, Hilton, Kelton, Miller, Simens, Askell, Welinder, Christiano, Leike, and Lowe]{ouyang2022traininglanguagemodelsfollow}
Long Ouyang, Jeff Wu, Xu~Jiang, Diogo Almeida, Carroll~L. Wainwright, Pamela Mishkin, Chong Zhang, Sandhini Agarwal, Katarina Slama, Alex Ray, John Schulman, Jacob Hilton, Fraser Kelton, Luke Miller, Maddie Simens, Amanda Askell, Peter Welinder, Paul Christiano, Jan Leike, and Ryan Lowe.
\newblock Training language models to follow instructions with human feedback, 2022.
\newblock URL \url{https://arxiv.org/abs/2203.02155}.

\bibitem[Qwen et~al.(2025)Qwen, Yang, Yang, Zhang, Hui, Zheng, Yu, Li, Liu, Huang, Wei, Lin, Yang, Tu, Zhang, Yang, Yang, Zhou, Lin, Dang, Lu, Bao, Yang, Yu, Li, Xue, Zhang, Zhu, Men, Lin, Li, Tang, Xia, Ren, Ren, Fan, Su, Zhang, Wan, Liu, Cui, Zhang, and Qiu]{qwen_qwen25_2025}
Qwen, An~Yang, Baosong Yang, Beichen Zhang, Binyuan Hui, Bo~Zheng, Bowen Yu, Chengyuan Li, Dayiheng Liu, Fei Huang, Haoran Wei, Huan Lin, Jian Yang, Jianhong Tu, Jianwei Zhang, Jianxin Yang, Jiaxi Yang, Jingren Zhou, Junyang Lin, Kai Dang, Keming Lu, Keqin Bao, Kexin Yang, Le~Yu, Mei Li, Mingfeng Xue, Pei Zhang, Qin Zhu, Rui Men, Runji Lin, Tianhao Li, Tianyi Tang, Tingyu Xia, Xingzhang Ren, Xuancheng Ren, Yang Fan, Yang Su, Yichang Zhang, Yu~Wan, Yuqiong Liu, Zeyu Cui, Zhenru Zhang, and Zihan Qiu.
\newblock Qwen2.5 {Technical} {Report}, January 2025.
\newblock URL \url{http://arxiv.org/abs/2412.15115}.
\newblock arXiv:2412.15115 [cs].

\bibitem[Mercer et~al.(2025)Mercer, Spillard, and Martin]{mercer2025briefanalysisDeepSeekr1}
Sarah Mercer, Samuel Spillard, and Daniel~P. Martin.
\newblock Brief analysis of deepseek r1 and its implications for generative ai, 2025.
\newblock URL \url{https://arxiv.org/abs/2502.02523}.

\bibitem[Shao et~al.(2024)Shao, Wang, Zhu, Xu, Song, Bi, Zhang, Zhang, Li, Wu, and Guo]{shao_deepseekmath_2024}
Zhihong Shao, Peiyi Wang, Qihao Zhu, Runxin Xu, Junxiao Song, Xiao Bi, Haowei Zhang, Mingchuan Zhang, Y.~K. Li, Y.~Wu, and Daya Guo.
\newblock {DeepSeekMath}: {Pushing} the {Limits} of {Mathematical} {Reasoning} in {Open} {Language} {Models}, April 2024.
\newblock URL \url{http://arxiv.org/abs/2402.03300}.
\newblock arXiv:2402.03300 [cs].

\bibitem[Pan et~al.(2025)Pan, Zhang, Wang, Yuan, Peng, and Suhr]{pan2025tinyzero}
Jiayi Pan, Junjie Zhang, Xingyao Wang, Lifan Yuan, Hao Peng, and Alane Suhr.
\newblock Tinyzero.
\newblock \url{https://github.com/Jiayi-Pan/TinyZero}, 2025.

\bibitem[Yuan et~al.(2025)Yuan, Pang, Cho, Li, Sukhbaatar, Xu, and Weston]{yuan_self-rewarding_2025}
Weizhe Yuan, Richard~Yuanzhe Pang, Kyunghyun Cho, Xian Li, Sainbayar Sukhbaatar, Jing Xu, and Jason Weston.
\newblock Self-{Rewarding} {Language} {Models}, March 2025.
\newblock URL \url{http://arxiv.org/abs/2401.10020}.
\newblock arXiv:2401.10020 [cs].

\bibitem[Simonds and Yoshiyama(2025)]{simonds_ladder_2025}
Toby Simonds and Akira Yoshiyama.
\newblock {LADDER}: {Self}-{Improving} {LLMs} {Through} {Recursive} {Problem} {Decomposition}, March 2025.
\newblock URL \url{http://arxiv.org/abs/2503.00735}.
\newblock arXiv:2503.00735 [cs].

\bibitem[Burns et~al.()Burns, Izmailov, Kirchner, Baker, Gao, Aschenbrenner, Chen, Ecoffet, Joglekar, Leike, Sutskever, and Wu]{burns_weak--strong_nodate}
Collin Burns, Pavel Izmailov, Jan~Hendrik Kirchner, Bowen Baker, Leo Gao, Leopold Aschenbrenner, Yining Chen, Adrien Ecoffet, Manas Joglekar, Jan Leike, Ilya Sutskever, and Jeff Wu.
\newblock {WEAK}-{TO}-{STRONG} {GENERALIZATION}: {ELICITING} {STRONG} {CAPABILITIES} {WITH} {WEAK} {SUPERVISION}.

\bibitem[Christiano et~al.(2018)Christiano, Shlegeris, and Amodei]{christiano_supervising_2018}
Paul Christiano, Buck Shlegeris, and Dario Amodei.
\newblock Supervising strong learners by amplifying weak experts, October 2018.
\newblock URL \url{http://arxiv.org/abs/1810.08575}.
\newblock arXiv:1810.08575 [cs].

\bibitem[Hubinger et~al.(2021)Hubinger, Merwijk, Mikulik, Skalse, and Garrabrant]{hubinger_risks_2021}
Evan Hubinger, Chris~van Merwijk, Vladimir Mikulik, Joar Skalse, and Scott Garrabrant.
\newblock Risks from {Learned} {Optimization} in {Advanced} {Machine} {Learning} {Systems}, December 2021.
\newblock URL \url{http://arxiv.org/abs/1906.01820}.
\newblock arXiv:1906.01820 [cs].

\bibitem[Le et~al.(2022)Le, Wang, Gotmare, Savarese, and Hoi]{le_coderl_2022}
Hung Le, Yue Wang, Akhilesh~Deepak Gotmare, Silvio Savarese, and Steven C.~H. Hoi.
\newblock {CodeRL}: {Mastering} {Code} {Generation} through {Pretrained} {Models} and {Deep} {Reinforcement} {Learning}, November 2022.
\newblock URL \url{http://arxiv.org/abs/2207.01780}.
\newblock arXiv:2207.01780 [cs].

\end{thebibliography}





\end{document}